\documentclass[a4paper,conference]{IEEEtran}
\ifCLASSINFOpdf
\else
\fi
\usepackage{booktabs}
\usepackage{xcolor}
\usepackage{multirow}
\usepackage{subcaption}
\usepackage{times}
\usepackage{epsfig}
\usepackage{graphicx}
\usepackage{amsmath}
\usepackage{amssymb}
\usepackage[linesnumbered,ruled,vlined]{algorithm2e}

\SetCommentSty{mycommfont}
\usepackage{float}

\begin{document}
%

\title{Pixel to Binary Embedding Towards Robustness for CNNs}

\author{\IEEEauthorblockN{Ikki Kishida and Hideki Nakayama}
\IEEEauthorblockA{The University of Tokyo\\
7-3-1 Hongo Bunkyo-ku, Tokyo, Japan\\
kishida@nlab.ci.i.u-tokyo.ac.jp and nakayama@ci.i.u-tokyo.ac.jp}
}


%


\maketitle

\begin{abstract}
There are several problems with the robustness of Convolutional Neural Networks (CNNs).
For example, the prediction of CNNs can be changed by adding a small magnitude of noise to an input, and the performances of CNNs are degraded when the distribution of input is shifted by a transformation never seen during training (e.g., the blur effect).
There are approaches to replace pixel values with binary embeddings to tackle the problem of adversarial perturbations, which successfully improve robustness.
In this work, we propose Pixel to Binary Embedding (P2BE) to improve the robustness of CNNs.
P2BE is a learnable binary embedding method as opposed to previous hand-coded binary embedding methods.
P2BE outperforms other binary embedding methods in robustness against adversarial perturbations and visual corruptions that are not shown during training.
\end{abstract}

\IEEEpeerreviewmaketitle

\section{Introduction}

Convolutional Neural Networks (CNNs) have several issues with robustness.
One of the problems is adversarial perturbations: they can maliciously modify CNN's prediction by adding a small magnitude of noise to an input \cite{adv_intriguing}.
Since the finding of adversarial perturbations, many types of attacking methods \cite{adv_explain,adv_pgd} and defensive methods \cite{adv_quantization,adv_randomresizepad} have been proposed.
We also know that CNNs do not generalize on input distributions other than the one they are trained on \cite{distortions}.
For example, CNNs trained with regular images fail to generalize on images with the blur effect \cite{blur}.
CIFAR-C and ImageNet-C \cite{datasetc} are proposed to investigate the generalization ability of trained models on such visually corrupted images.
Since then, some training strategies \cite{augmix,robust_simple} and ensemble techniques \cite{robust_ensemble} have been proposed to improve the robustness against visual corruptions which do not appear during training.
Robustness matters for applying the computer vision system to real-world applications since some malicious exploitations may occur using the above flaws.

For the robustness against adversarial perturbations, there are approaches to replace pixel values with binary embeddings (e.g., one-hot and thermometer encoding \cite{adv_thermometer}).
They empirically show that binarized input successfully improves the robustness against adversarial perturbations.
These binary encodings are based on hand-coded simple rules even though vision tasks are diverse and complex.
We consider that such a simple binary encoding would not be optimal for all problems. It is a promising direction to learn the rule of binary encoding for each problem by using data.

In this work, we propose Pixel to Binary Embedding (P2BE), which is a learnable binary embedding method as opposed to previous hand-coded binary embeddings \cite{adv_thermometer}.
In addition, we propose embedding smoothness loss to introduce the effect of quantization which effectively works with adversarial perturbations.

To measure the robustness against visual corruptions, we use two benchmark datasets for image classification (i.e., CIFAR-100-C and ImageNet-C datasets).
We show that P2BE outperforms other binary encoding methods for robustness against visual corruptions across various CNNs.
In addition, we show that P2BE achieves the best robustness performances against adversarial perturbations.
In our analyses, we show that the performance of P2BE is not sensitive to the dimension size $M$, the proposed embedding smoothness loss is essential to improve robustness against adversarial perturbations, and ImageNet-1k pretrained P2BE has the transferability to the other task. 

\begin{figure}[t]
    \centering
    \includegraphics[width=0.49\textwidth]{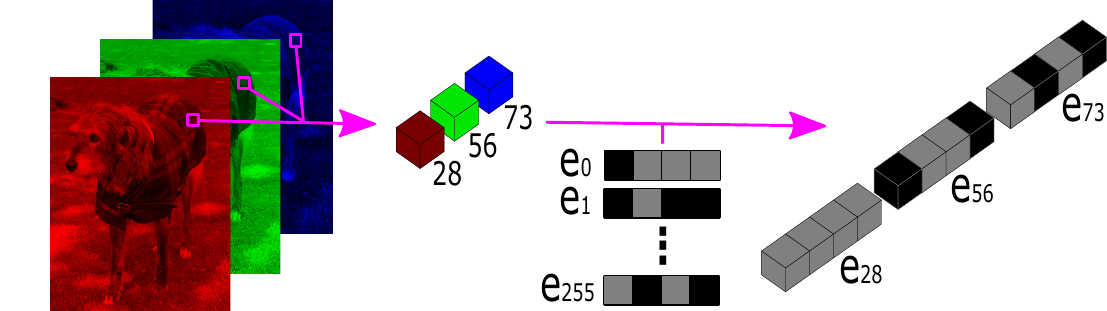}
    \caption{Overview of P2BE (Pixel to Binary Embedding). Each value of RGB image have 256 types of magnitude (i.e., 0 to 255) and P2BE replaces each RGB value with learned binary embedding $e_k \in \{0,1\}^{M}$ where $k \in [0,...,255]$. $M \in \mathbb{N}$ is the dimension of the embedding and hyperparameter controlling the expression ability of $e_k$. The black and grey colors indicate 0 and 1, respectively. The figure illustrates the case of $M=4$.}
    \label{fig:overview}
\end{figure}

Our contributions are summarized as follows:
\begin{itemize}
    \item We propose P2BE, which is a learnable binary embedding unlike other hand-coded binary embedding methods. P2BE shows the best robustness performances among RGB and other binary embedding methods on various datasets.
    \item The embedding smoothness loss is proposed to realize the effect of quantization in P2BE. Our analysis shows that the embedding smoothness loss improves robustness against adversarial perturbations in P2BE.
    \item Binary embedding methods have been evaluated only from the view of robustness against adversarial perturbations. In this work, we additionally assess binary embedding methods on robustness against visual corruptions. The results reveal that the approach of binary embedding effectively improves robustness against visual corruptions.
\end{itemize}

\section{Related Work}

\subsection{Robustness of CNNs}

\noindent \textbf{Adversarial Perturbations.}
A small amount of adversarial noise can intentionally change the prediction of trained CNNs.
This phenomenon is called adversarial perturbations, and it was initially reported in \cite{adv_intriguing}.
Since the finding of adversarial perturbations, many types of attacking methods \cite{adv_explain,adv_pgd,adv_deepfool,adv_cw,adv_universal,adv_yopo,adv_onepixel} and defensive methods \cite{adv_quantization,adv_thermometer,adv_localintrinsicdimensionality,adv_tvminimization,adv_stochasticactivation,adv_randomresizepad,adv_pixeldefend,adv_rbs,adv_obfuscatedgradients,adv_selfensemble,adv_bitdepth,adv_random} have been proposed.
However, it has been reported that the most robust defensive method is still defective against some attacks \cite{adv_nipscomp}.

Some defensive methods improve robustness against adversarial perturbations by transforming the input $x \in [0, 1, \cdots, 255]$ into an $M$-dimensional binary embedding.
One-hot encoding $D_{\text{one-hot}}(x) \in \{0,1\}^{M}$ is a simple binary discretization method \cite{adv_thermometer} as follows:

\begin{equation}
D_{\text{one-hot}}(x)_{i} = 
    \begin{cases}
    1,& \text{if } \frac{i-1}{M} \leq \frac{x}{255} < \frac{i}{M}  \text{ or } x = 255\\
    0,              & \text{otherwise,}
\end{cases}
\label{eq:onehot}
\end{equation}
where $i \in \{1,...,M\}$, and $M \in \mathbb{N}^{+}$ is a hyperparameter controling the size of the dimension of the binary embedding.
$D_{\text{one-hot}}$ improves the robustness of CNNs, however, it degrades the performance on clean images by losing the information on relative distance (e.g., $D_{\text{one-hot}}(0.03)$ is equally far from $D_{\text{one-hot}}(0.48)$ and $D_{\text{one-hot}}(0.92)$).
To overcome such flaw, thermometer encoding $D_{\text{thermo}}(x) \in \{0, 1\}^{M}$ \cite{adv_thermometer} is proposed as follows:
\begin{equation}
D_{\text{thermo}}(x)_{i} = 
    \begin{cases}
    1,& \text{if } \frac{x}{255} < \frac{i}{M} \text{ or } x = 255 \\
    0,              & \text{otherwise.}
\end{cases}
\label{eq:thermo}
\end{equation}
The examples of binary encoding methods are summarized in Table \ref{tab:properties}.

\noindent \textbf{Common Visual Corruptions.}
CNNs fail to generalize on the images with visual corruptions, which are not shown during training \cite{distortions,blur}.
It is essential to measure the robustness against such common visual corruptions (e.g., blur, brightness, contrast, and so on) for the reliability of computer vision systems.
For benchmarking the robustness of image classification, ImageNet-C and CIFAR-C \cite{datasetc} are proposed, and there are 15 types of common visual corruptions for evaluation.
PASCAL-C, COCO-C, and Cityscapes-C are proposed for evaluating the robustness of object detection \cite{detection_c} by using the same types of common visual corruptions.

\noindent \textbf{Robust Training.}
Some training methods are proposed to improve the robustness of a vision system against adversarial perturbations or visual corruptions.
\cite{adv_cw} proposed Adversarial Training (AT) and they train neural networks on only adversarial perturbed images.
AT improves the performance on adversarial perturbed images in exchange for dropping the performance on clean images \cite{adv_tradeoff,adv_tradeoff2}.

Several approaches aim to improve robustness against visual corruptions which never seen during training.
One of the approaches is \texttt{augxmix} training method \cite{augmix}. \texttt{augxmix} introduces the regularization loss $L_{\text{aug}}$ to enforce the model to do consistent predictions between the original and visually transformed images.
$L_{\text{aug}}$ is computed based on Jensen-Shannon divergence (JSD) between original image (i.e., $x$) and two visually transformed images (i.e., $x_{\text{aug1}}$ and $x_{\text{aug2}}$) as follows:
\begin{equation}
\begin{split}
& L_{\text{aug}}(p(x); p(x_{\text{aug1}}); p(x_{\text{aug2}}))= \\ &\frac{1}{3} (\text{KL}\left [p(x) \middle\|  V \right] + \text{KL}\left [p(x_{\text{aug1}}) \middle\|  V \right ] + \text{KL}\left [p(x_{\text{aug2}}) \middle\|  V \right]),
\end{split}
\end{equation}
where $V$ is $\frac{1}{3}(p(x) + p(x_{\text{aug1}}) + p(x_{\text{aug2}}))$, $\text{KL}$ is Kullback–Leibler divergence and $p$ is CNN's prediction from the softmax layer.
Another approach is an adversarial training method to use an adversarial noise generator \cite{robust_simple}.
They show that being robust against noise improves the robustness against common visual corruptions.

\begin{table}[t]
  \centering
  \caption{Examples of binary embedding methods. $M$ is the dimensional size of embeddings. The examples in the table represent the case of $M=10$. P2BE is a learnable binary embedding method. Thus the binary translation rules of P2BE depend on tasks and training strategies.}
 \label{tab:properties}
  \vspace{0.1cm}
  \setlength{\tabcolsep}{2.0pt}
  \begin{tabular}{ccccc}
 & & One-hot \cite{adv_thermometer} & Thermometer \cite{adv_thermometer} &     P2BE (ours)\\
  \hline\noalign{\smallskip}
  & 0.03  & \texttt{\footnotesize{[1000000000]}} & \texttt{\footnotesize{[1111111111}]}& \texttt{\footnotesize{[0111101011]}} \\
  Values & 0.48  & \texttt{\footnotesize{[0000100000]}} & \texttt{\footnotesize{[0000111111]}} & \texttt{\footnotesize{[1111101001]}} \\
  & 0.92 & \texttt{\footnotesize{[0000000001]}} & \texttt{\footnotesize{[0000000001]}} & \texttt{\footnotesize{[1011011110]}} \\
  \hline
  \end{tabular}
\end{table}

\subsection{Binary Neural Netowrks}

Deep Neural Networks (DNNs) require heavy matrix computations. Therefore, it is hard to deploy DNNs on devices that have limited computational ability.
To overcome such issues, Binary Neural Networks (BNNs) have been proposed \cite{bnn_clip,bnn_xnor,bnn_binaryconnect,bnn_dorefa,bnn_lqnet,bnn_approxsign,bnn_softquantization} .
In BNNs, the matrix multiplications are replaced with the combinations of bitwise XNOR and bit count operations, which are lightweight calculations.
BNNs accelerate inference time, save up storage, and improve energy efficiency. However, BNNs suffer from performance degradation compared to non-binary DNNs.
The motivation of BNNs and binary embedding methods are different.
On the one hand, BNNs aim to make DNNs resource-efficient.
On the other hand, binary embedding methods translate only input into binary values for improving robustness.

\section{Method}

\subsection{Pixel to Binary Embedding}

We show the overview of P2BE in Fig \ref{fig:overview}.
Our method transforms an RGB image $x \in \{0,...,255\}^{3\times H \times W}$ to the learnable binary embedding $b \in \{0,1\}^{3M \times H \times W}$ where $M \in \mathbb{N^{+}}$ is the dimension size of the binary embedding.
There are two steps in P2BE: Binarization and Embedding Smoothness Loss.

\noindent \textbf{Binarization.} In P2BE, the learnable embeddings $w_k \in \mathbb{R}^{M}$ are translated into the binary embedding $e_k \in \{0,1\}^{M}$ where $k \in [0,...,255]$ corresponds to the magnitude of each RGB value and $M \in \mathbb{N}^{+}$ is the hyperparameter to controlling the dimension size of $e_k$.
The binarization is based on the \texttt{sign} function as follows:
\begin{equation}
\texttt{sign}(x) = 
    \begin{cases}
    1,& \text{if } x \geq 0\\
    -1,              & \text{otherwise.}
    \end{cases}
\end{equation}
In P2BE, we calculate the binary embedding $e_k$ as follows:
\begin{equation}
e_k = 0.5 \times \texttt{sign}(w_k) + 0.5.
\label{eq:sign}
\end{equation}

Since the \texttt{sign} function is non-differentiable, straight-through estimator (STE) has been proposed to make it differentiable \cite{bnn_ste}.
STE approximates $\frac{\delta \texttt{sign}(x)}{\delta x}$ by the derivative of the \texttt{identity} function.
However, using $\frac{\delta \texttt{identity}(x)}{\delta x}$ as backward function leads unstablities in learning since the \texttt{identity} and \texttt{sign} functions are greatly different.
Since then, differentiable functions closer to the \texttt{sign} function have been proposed for better approximation of $\frac{\delta \texttt{sign}(x)}{\delta x}$ \cite{bnn_clip,bnn_approxsign,bnn_softquantization,bnn_bnnplus,bnn_circulant,bnn_binaryduo}.

In P2BE, we use a function called approximate sign (i.e., $\texttt{sign}_{\texttt{approx}}$) \cite{bnn_approxsign} and its derivatives are as follows:

\begin{equation}
\frac{\delta \texttt{sign}_{\texttt{approx}}(x)}{\delta x} = 
\begin{cases}
    2 + 2x,& \text{if } -1 \leq x < 0\\
    2 - 2x,& \text{if } 0 \leq x < 1\\
    0,              & \text{otherwise.}
\end{cases}
\end{equation}
$\texttt{sign}_{\texttt{approx}}$ function approximates \texttt{sign} function with quadratic equation.
When we calculate the gradients of Eq \ref{eq:sign} to $w_k$, we use $\frac{\delta \texttt{sign}_{\texttt{approx}}(w_k)}{\delta w_k}$ instead of $\frac{\texttt{sign}(w_k)}{w_k}$.
 P2BE transforms the each RGB value $x_{c,h,w} \in [0,...,255]$ to binary values as follows:
\begin{equation}
    D_{\text{p2be}}(x_{c,h,w}) = e_{x_{c,h,w}},
\end{equation}
where $c$, $h$ and $w$ are the channel, height and width of an input, respectively.
The pseudocode of P2BE is shown in Algorithm \ref{algorithm_p2be}.

\noindent \textbf{Embedding Smoothness Loss.}
As \cite{adv_explain} have hypothesized, the adversarial perturbations are caused by the linearity of trained neural networks with respect to an input.
Let us consider the case of the single linear layer with sigmoid function $\sigma$.
\begin{equation}
    y(\hat{x}) = \sigma (w(x + \epsilon)) = \sigma (w \cdot x + w \cdot \epsilon),
    \label{linear}
\end{equation}
where $w \in \mathbb{R}^{m \times n}$, $x \in \mathbb{R}^{m}$ and $\epsilon \in \mathbb{R}^{m}$ are the weight, input and adversarial noise, respectively.
$\epsilon$ satisfies $\|\epsilon\|_{\infty} \le C$ where $C \in \mathbb{R}^{+}$ is small enough.
If the dimension $m$ is large enough, the output of the sigmoid function can be changed significantly by $w \cdot \epsilon$.

As \cite{adv_thermometer} have mentioned, the quantization can be a reasonable approach against adversarial perturbations since quantized $D_{\text{one-hot}}(x)$ may be equivalent to quantized $D_{\text{one-hot}}(\hat{x})$ and the term of $w \cdot \epsilon$ disappears.
As can be seen in Eq \ref{eq:onehot} and \ref{eq:thermo}, the quantization of $D_{\text{one-hot}}$ and $D_{\text{thermo}}$ \cite{adv_thermometer} are pre-defined (e.g., location and the step size of quantization).
We propose the embedding smoothness loss to introduce the effect of quantization in P2BE.
The embedding smoothness loss $L_{\text{smooth}}$ is computed by cosine similarity $\cos(a\angle b)$: $\mathbb{R}^{d} \times \mathbb{R}^{d} \rightarrow [1,-1]$ as follows:
\begin{equation}
\begin{split}
    L_{\text{smooth}}(w) &= \sum_{k \in \left [0, ..., 254 \right ]} 1 - \cos(w_k \angle w_{k+1})\\ 
    &= \sum_{k \in \left [0, ..., 254 \right ]} 1 - \frac{<w_k, w_{k+1}>}{\|w_k\| \|w_{k+1}\|},
\end{split}
\end{equation}
where $\verb|<|\cdot,\cdot \verb|>|$ denotes the dot product and $\| \cdot \|$ represents the l2 norm.
The smaller $L_{\text{smooth}}$, the angle of neighbored embeddings $w_k$ and $w_{k+1}$ is closer to 0.

\begin{algorithm}[t]

\SetKwInOut{Input}{Input}
\SetKwInOut{Output}{Output}
\DontPrintSemicolon
\SetAlgoLined
\SetNoFillComment
\LinesNotNumbered 

\Input{Image $x \in \{0,1,\cdots 255\}^{3 \times K \times N}$, \\Learnable Embedding $w \in \mathbb{R}^{256 \times M}$, \\ Loss $L$}
\textbf{Initialization:} $w \sim \mathcal{N}(0,1)
$\\
\textbf{Forwarding $(x)$:}\\
\tcp*[l]{$e  \in \{0,1\}^{256\times M}$}
$e = \frac{\texttt{sign}(w)+1}{2}$\\
 \For{$c = 0, \cdots , 2$}{
    \For{$k = 0, \cdots , K - 1$}{
        \For{$n = 0, \cdots, N - 1$}{
                \For{$m = 0, \cdots, M - 1$}{
\tcp*[l]{$b_{Mc+m,k,n} \in \{0,1\}$}
        $b_{Mc+m,k,n} = e_{x_{c,k,n},m}$
        }
        }
    }
}
\tcp*[l]{$b \in \{0,1\}^{MC\times K \times N}$}
return $b$\\

\textbf{Backwarding $(\frac{\delta L}{\delta b})$}:\\
\tcp*[l]{$\frac{\delta L}{\delta b} \in \mathbb{R}^{3M \times K \times N}$}
 \For{$c = 0, \cdots , 2$}{
    \For{$k = 0, \cdots , K - 1$}{
        \For{$n = 0, \cdots, N - 1$}{
                \For{$m = 0, \cdots, M - 1$}{
\tcp*[l]{$\times$ is scalar multiplication}
\tcp*[l]{$\frac{\delta L}{\delta w_{x_{c,k,n},m}} \in \mathbb{R}$}
$\frac{\delta L}{\delta w_{x_{c,k,n},m}}  \mathrel{{+}{=}}$\\
\small{$\frac{1}{2}(\frac{\delta L}{\delta b_{Mc+m,k,n}} \times \frac{\delta \text{\texttt{sign}}_{\text{\texttt{approx}}}(w_{x_{c,k,n},m})}{\delta w_{x_{c,k,n},m}})$}\\
           }
        }
    }
}
\tcp*[l]{$\frac{\delta L}{\delta w} \in \mathbb{R}^{256 \times M}$}
return $\frac{\delta L}{\delta w}$
    \caption{P2BE Pseudocode}
    \label{algorithm_p2be}
\end{algorithm}

\section{Experiments}

\subsection{Preparations}
\label{preparations}

\noindent \textbf{Dataset.}
We use CIFAR-100 \cite{cifar} and ImageNet-1k \cite{ILSVRC} for our experiments.
CIFAR-100 are image classification datasets with 100 classes. They contain 50000 training images and 10000 validation images.
We use three types of models: Wide ResNet 40-2 \cite{wideresnet}, DenseNet-BC ($k$=12, $d$=100) \cite{densenet} and ResNeXt-29 ($32 \times 4$) \cite{resnext}.

ImageNet-1k is the large-scale dataset for image classification with 1.28M training images and 50k validation images of 1000 classes.
In this work, we use ResNet50 \cite{resnet} as the baseline model for ImageNet-1k experiments.

\noindent \textbf{Evaluation.}
We evaluate the robustness of models in two aspects: common visual corruptions and adversarial perturbations.
As the benchmarks of the robustness against common visual corruptions, we use CIFAR-100-C and ImageNet-C datasets \cite{datasetc}.
Fifteen types of visual corruptions $c$ transform the images with five different severities $s$ (e.g., blurring, contrasting).
On CIFAR-100-C, we calculate the average of the test error $\text{E}_{s,c}$ across all corruptions $c$ and severities $s$.
On ImageNet-C, we calculate the mean Corruption Error (mCE) for the measurement of the robustness as proposed in \cite{datasetc}.
mCE is the average of Corruption Error ($\text{CE}_c$) across all corruptions $c$. $\text{CE}_{c}$ is normalized test error as follows: 
$\text{CE}_{c} = \sum^{5}_{s=1} \text{E}_{s,c} / \sum^{5}_{s=1} \text{E}_{s,c}^{\text{alexnet}}$ where $\text{E}_{s,c}^{\text{alexnet}}$ is the test error of alexnet \cite{alexnet}.

To evaluate robustness against adversarial perturbations, we measure the test error on adversarially perturbed test images.
Since binary embedding methods are not differentiable, ordinal attacking methods of adversarial perturbations are not applicable.
Thus, we generate adversarial noise for testing by using the LS-PGA attacking method \cite{adv_thermometer}, which is specifically designed for the network with binary embeddings.

\subsection{Common Visual Corruptions}
\label{sec:impl}
\noindent \textbf{Implementation Details.} As the optimizer for the classification models, we use Momentum SGD with momentum of 0.9.
On the CIFAR-100 dataset, we train the models for 200 epochs with 128 batches. We set the coefficient of weight decay to $5.0 \times 10 ^{-4}$, and the learning rate is scheduled by cosine annealing strategy, which starts from 0.1 and ends at $1.0 \times 10 ^{-5}$.
On the ImageNet-1k dataset, we train the models for 180 epochs with 64 batches.
The coefficient of weight decay is $1.0 \times 10 ^{-4}$ and the initial learning rate is set to 0.1 and divided by 10 at 60 and 120 epochs.

As the optimizer for embedding parameters of P2BE, we use AdamW \cite{adamw}.
The learning rate, $\beta_1$ and $\beta_2$ are set to $1.0 \times 10 ^{-4}$, 0.999 and 0.999, respectively.
We do not apply the scheduling of the learning rate, and the coefficient of weight decay is set to $1.0 \times 10 ^{-4}$.
The dimension of binary embedding $M$ is 64.

The total loss is defined as follows:
\begin{equation}
    L_{\text{total}} = L_{\text{ce}} + \alpha L_{\text{aug}} + \lambda L_{\text{smooth}},
    \label{eq_c}
\end{equation}
where $L_\text{ce}$ is cross-entropy loss for the classification and $\alpha \in \mathbb{R}^{+}$ is the hyperparameter for \texttt{augmix} regularization loss.
$\lambda \in \mathbb{R}^{+}$ is the hyperparameter controling the degree of quantization.
When the $\lambda$ is larger, the neighbored embeddings (i.e., $w_k$ and $w_{k+1}$) tend to have similar directions. 
For the training of P2NE, the coefficients $\lambda$ are set to 1.0, and 10.0 on CIFAR-100-C and ImageNet-C, respectively.
$\alpha$ is set to 12 which is the same value used in \cite{augmix}. 

\noindent \textbf{Results.}
The result of CIFAR-100-C is shown in Table \ref{tab:cifar_c}.
It shows that the approaches of binary embedding generally improve the robustness against visual corruptions with small performance drops on clean images compared to RGB input space.
This finding is interesting since the binary embeddings have only been evaluated from the aspect of robustness against adversarial perturbations.
It implies that designing a sophisticated input space may be a promising way to improve robustness against never-seen visual corruptions.

As can be seen in Table \ref{tab:cifar_c}, one-hot encoding has the bigger performance drops on clean images and bigger improvements in the robustness against never-seen visual corruptions among the three binary embedding methods.
Thermometer encoding has the opposite tendencies: the smaller performance drops on clean images and the smaller robustness improvement against never-seen visual corruptions.
P2BE has the good properties of both methods that smaller performance drops on clean images and a bigger improvement of robustness against never-seen visual corruptions.

The result of ImageNet-C is shown in Table \ref{tab:imagenet_c}.
The results are similar to those in CIFAR-100-C datasets and P2BE shows the best robustness against never-seen visual corruptions.
As can be seen in Table \ref{tab:imagenet_c}, the binary embedding methods tend to improve robustness against corruptions of the noise and digital categories. However, it has almost no effect on the weather category.
This result indicates the limitation of the approach of binary embeddings.

    \begin{table}[tb]
      \centering
      \caption{The test error of CIFAR-100-C. The all models are trained with Eqn. \ref{eq_c}. $\lambda$ is set to 0 in the case of RGB, One-hot and Thermometer. The values on the table represent mean $\pm$ std across 5 runs. The values in parenthesis are the test error of CIFAR-100 (i.e., clean images).}
      \label{tab:cifar_c}
      \begin{tabular}{cccccc}
      \hline\noalign{\smallskip}
      \normalsize{Encoding} & \normalsize{Model} & \normalsize{Test error (\%)}\\
      \hline  \noalign{\smallskip}
      & \normalsize{WideResNet}& 35.2$\pm$0.3 (22.2$\pm$0.3) \\
      RGB & \normalsize{DenseNet}&37.3$\pm$0.1 (22.6$\pm$0.3) \\
      & \normalsize{ResNeXt} & 33.9$\pm$0.3 (20.5$\pm$0.4)\\

      \hline \noalign{\smallskip}
       & \normalsize{WideResNet}& 34.4$\pm$0.4 (23.9$\pm$0.2)\\
      One-hot & \normalsize{DenseNet}& 36.6$\pm$0.2 (23.9$\pm$0.2)\\
      & \normalsize{ResNeXt} & 33.2$\pm$0.3 (21.8$\pm$0.2)\\

      \hline \noalign{\smallskip}
            & \normalsize{WideResNet}& 35.1$\pm$0.2 (22.8$\pm$0.2) \\
      Thermometer & \normalsize{DenseNet}& 36.9$\pm$0.2 (23.0$\pm$0.1) \\
      & \normalsize{ResNeXt} & 33.9$\pm$0.3 (21.3$\pm$0.3)\\
      \hline \noalign{\smallskip}
      & \normalsize{WideResNet}&  {\textbf{34.2$\pm$0.2} (22.8$\pm$0.3)}\\
      P2BE (ours) & \normalsize{DenseNet}& {\textbf{36.3 $\pm$0.2} (23.3$\pm$0.3)}\\
      & \normalsize{ResNeXt} & {\textbf{32.6$\pm$0.2} (20.7$\pm$0.4)}\\
      \hline
      \end{tabular}
    
    \end{table}
 
\begin{table*}[tb]

  \caption{The clean error of ImageNet-1k and Corruption Error ($\text{CE}_c$) of ImageNet-C. $\text{CE}_c$ is the normalized test error by the test error of alexnet. mCE is the averaged $\text{CE}_c$ across 15 different corruptions $c$. The detailed definition of mCE is denoted in Sec. \ref{preparations}. The all models are trained with Eqn. \ref{eq_c}. $\lambda$ is set to 0 in the case of RGB, One-hot and Thermometer. The lower the values on the table, the better performances.}
   \label{tab:imagenet_c}
 \setlength{\tabcolsep}{1.5pt}
  \centering
  \begin{tabular}{c|c|ccc|cccc|cccc|cccc|c}
     \multicolumn{1}{c}{} & \multicolumn{1}{c}{} &\multicolumn{3}{c}{\small{Noise}} & \multicolumn{4}{c}{\small{Blur}} & \multicolumn{4}{c}{\small{Weather}} & \multicolumn{4}{c}{\small{Digital}}\\
  \hline
     & \small{Clean} & \small{Gauss} & \small{Shot} & \small{Impulse} & \small{Defocus} & \small{Glass} & \small{Motion} & \small{Zoom} & \small{Snow} & \small{Frost} & \small{Fog} & \small{Bright} & \small{Contrast} & \small{Elastic} & \small{Pixel} & \small{Jpeg} & \small{mCE}\\
  \hline 
  \small{RGB} &  22.6 &     66     &    66  &     65   &    69  &  82    &      65  &     66      &     73   &    73  & 62  & 58   & 63 & 80 & 67 & 70 & 68.2 \\
  \small{One-hot}& 23.5  & 63 & 63 & 59 &  68  &  78    &  64   & 67   &73   &72    &65  & 59  & 64 & 74 & 56 & 68 & 66.1 \\
  \small{Thermometer}& 22.9 & 62 & 62  & 61 &    69  & 79 &  65  & 67  & 71   & 71  &  63 & 57 & 63 & 78 & 61  & 68 & 66.2 \\
  \small{P2BE (ours)}  &  23.3 &  62  &   61     &    61  &  67    &      77  &   63        & 64       &  71    & 71  &  63  & 58 & 63 & 74 & 61 & 67 & \textbf{65.6}\\
 \hline
  \end{tabular}
\end{table*}


\subsection{Adversarial Perturbations}

\begin{table}
  \centering
  \caption{The test error with LS-PGA white-box attack on CIFAR-10 and CIFAR-100. The values in parenthesis is the test error on clean images.}
  \label{tab:cifar_adv}
  \begin{tabular}{ccccccc}
  \hline\noalign{\smallskip}
  & & \footnotesize{One-hot}& \footnotesize{Thermometer}& \footnotesize{P2BE (ours)}\\
  \hline  \noalign{\smallskip}
\multirow{2}{*}{CIFAR-10} & \texttt{ConTrain} &66.1 (11.6)& 48.0 (16.0)& \textbf{45.3} (15.7)\\ 
                           & \texttt{AdvTrain} &52.1 (16.3) & 51.6 (15.0) & N/A \\
                          \cline{2-2} \noalign{\smallskip}
\multirow{2}{*}{CIFAR-100} & \texttt{ConTrain} &91.4 (38.1)  & 73.0 (42.7) & \textbf{71.5} (42.2)\\ 
                           & \texttt{AdvTrain} &75.2 (41.7) & 75.3 (41.0) & N/A \\
  \hline
  \end{tabular}
\end{table}

\noindent \textbf{Implementation Details.} The same hyperparameters for the training of CIFAR-100-C in Sec \ref{sec:impl} are used except for some specific parts in adversarial training.
Since binary embedding methods are not differentiable, ordinal attacking methods of adversarial perturbations are not applicable.
In this work, we use the LS-PGA attacking method \cite{adv_thermometer} which is specially developed to attack binary embeddings. 
For LS-PGA attacking, we use the seven steps for iterative attack with the annealing rate $\delta=1.2$ as \cite{adv_thermometer} use.
However, we notice that the step size of $\xi=0.031$ in \cite{adv_thermometer} is too small for the convergence on CIFAR-datasets.
In our experiments, we set $\xi$ to 1.0 for the convergence, and the results of one-hot and thermometer encoding are much worse than the scores reported in \cite{adv_thermometer}.

To be robust against adversarial perturbations, we use two types of adversarial training methods: \texttt{AdvTrain} and \texttt{ConTrain}. The standard adversarial training is to train the model on only adversarially perturbed images as proposed in \cite{adv_intriguing} and we call it \texttt{AdvTrain} in this work.
\texttt{ConTrain} is a variant of the \texttt{augmix} training method.
We use $L_\text{con}$ instead of $L_\text{aug}$ as follows:
\begin{equation}
\begin{split}
L_\text{con}(p(x); p({x_\text{adv}})) = \frac{1}{2} (\text{KL}\left [p(x) \middle\| V_{\text{con}}  \right] + \text{KL}\left [p(x_{\text{adv}}) \middle\| V_{\text{con}} \right]),
\end{split}
\end{equation}
where $x$ and $x_{\text{adv}}$ are the original and the adversarially perturbed images, respectively.
$V_{\text{con}}$ is $\frac{1}{2} (p(x) + p({x_\text{adv}}))$ and $p$ is the CNN's prediction from the softmax layer.

The total loss of \texttt{ConTrain} is defined as follows:
\begin{equation}
    L_{\text{total}} = L_{\text{ce}} + \alpha L_{\text{con}} + \lambda L_{\text{smooth}},
    \label{eq_adv}
\end{equation}
where the coefficients $\lambda$ are set to 1.0 and 0.1 on CIFAR-10 and CIFAR-100 datasets, respectively.

\noindent \textbf{Results.}
We show the results of adversarial perturbations in Table \ref{tab:cifar_adv}.
As can be seen in Table \ref{tab:cifar_adv}, P2BE with \texttt{ConTrain} achieves the best robustness against adversarial perturbations on both CIFAR-10 and CIFAR-100.
On one-hot and thermometer encoding, \texttt{AdvTrain} and \texttt{ConTrain} are suitable to improve the robustness against adversarial perturbations, respectively.
Interestingly, P2BE fails to learn with \texttt{AdvTrain}.
It seems P2BE requires non-perturbed images for the stability of learning.

\section{Analysis}

\noindent \textbf{Learned Binary Embedding in P2BE.}
\begin{figure*}[ht]
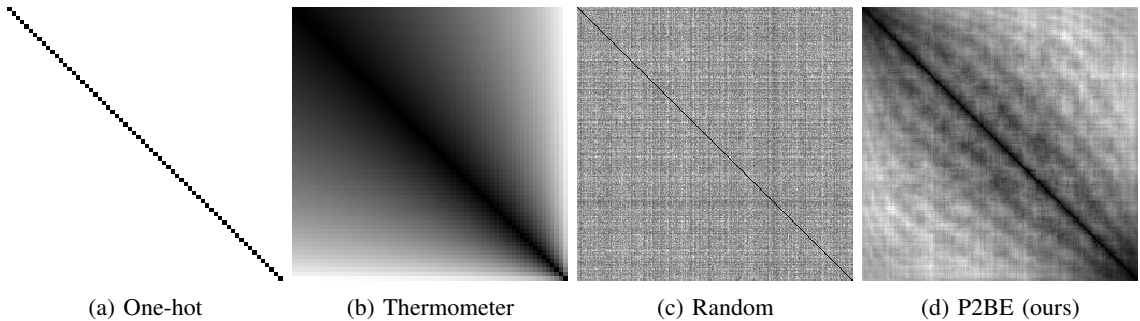

\centering
\begin{subfigure}{0.2\textwidth}
  \includegraphics[width=\linewidth]{figures/onehot.pdf}
  \caption{One-hot}
\end{subfigure}
\begin{subfigure}{0.2\textwidth}
  \includegraphics[width=\linewidth]{figures/thermo.pdf}
  \caption{Thermometer}
\end{subfigure}
\begin{subfigure}{0.2\textwidth}
  \includegraphics[width=\linewidth]{figures/random.pdf}
  \caption{Random}
\end{subfigure}
\begin{subfigure}{0.2\textwidth}
  \includegraphics[width=\linewidth]{figures/imagenet_p2be.pdf}
  \caption{P2BE (ours)}
\end{subfigure}
\caption{The cosine similarity of binary embeddings. The vertical and horizontal axis represents that the indices $i, j \in [0,...,255]$ corresponding to the magnitude for each RGB value. In table, the cell at the coordinate $(i,j)$ represents the cosine similarity between binary embeddings $e_i$ and $e_j$. The black and white colors indicate that the cosine similarities are 1.0 and 0.0, respectively. Figure (c) is the binary embedding generated by the standard normal distribution. Figure (d) is calculated by ImageNet-1k trained P2BE with ResNet50.}
\label{fig:cos}
\end{figure*}
We show the cosine similarities of binary embedding in Fig \ref{fig:cos}.
The distance space of ImageNet-1k trained P2BE is shown in Fig \ref{fig:cos}-(d).
It shows that the distances of P2BE embeddings are periodically larger and smaller.
Such properties of distance space do not exist in RGB, one-hot, and thermometer encoding.

\begin{figure}[btp]
    \centering
    \includegraphics[width=0.3\textwidth]{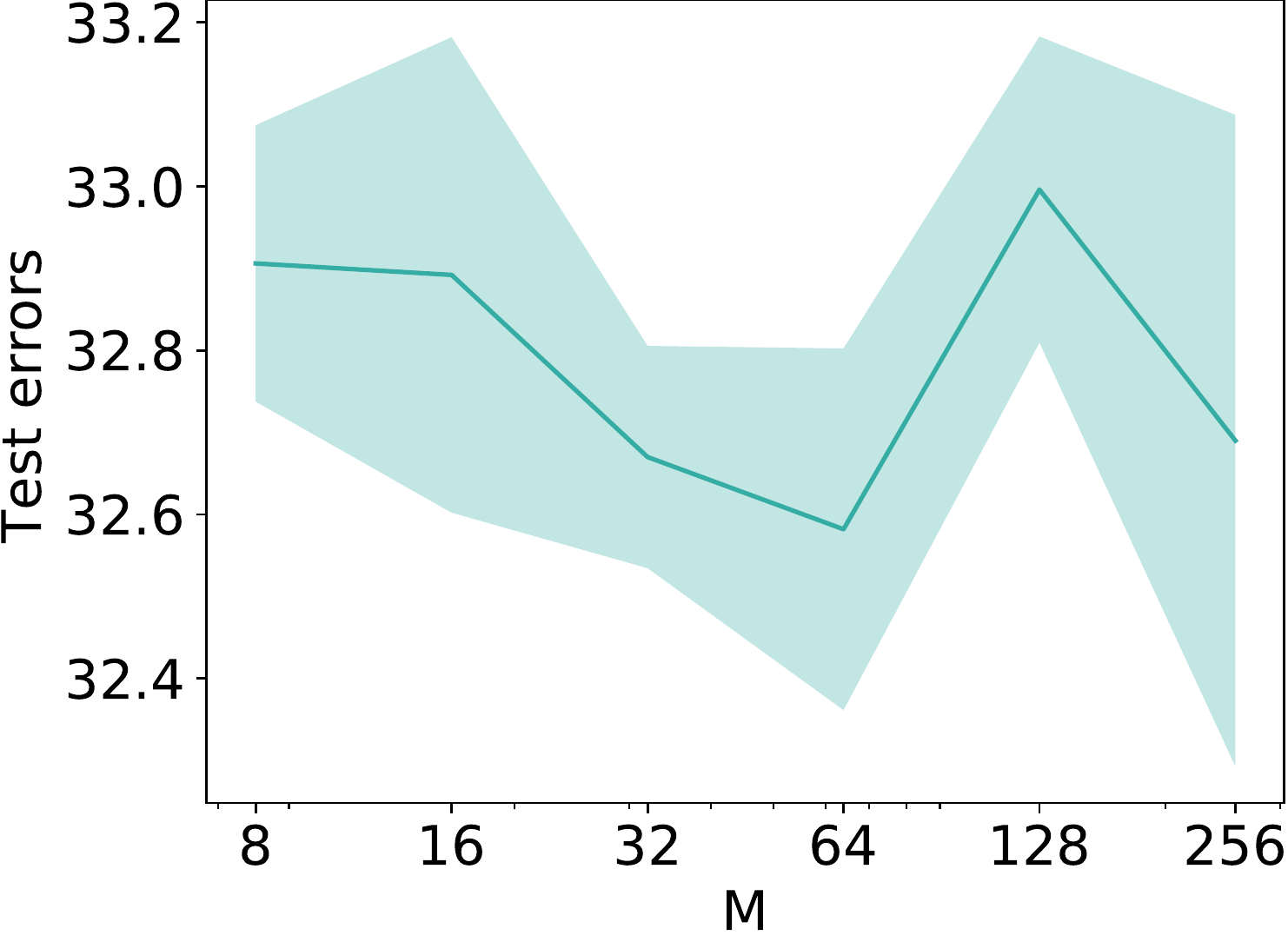}
    \caption{The CIFAR-100-C results of P2BE across various $M$. The horizontal axis is $M$ and the vertical axis is the test error of CIFAR-100-C.}
    \label{fig:ablation_m}
\end{figure}

\noindent \textbf{Effect of Dimension Size of Embedding M.}
We conduct an experiment to investigate the relationship between the robustness performance and dimension size $M$ of P2BE.
The result is shown in Fig \ref{fig:ablation_m}.
The worst performance is obtained when $M$ is 128.
As can be seen in Table \ref{tab:cifar_c}, it is still better than the results of other baselines. 
We claim that P2BE is not sensitive to the size of $M$ since the biggest performance gap is 0.5\% within various $M$.

\noindent \textbf{Effect of Embedding Smoothness Loss.}
We investigate the effectiveness of embedding smoothness loss.
Table \ref{tab:cifar_adv_lambda} shows the performances against adversarial perturbations on CIFAR-10 and CIFAR-100.
As can be seen in Table \ref{tab:cifar_adv_lambda}, P2BE with $\lambda=0.1$ outperforms P2BE with $\lambda=0.0$ which is the case without the embedding smoothness loss.
This result implies that embedding smoothness loss is an effective regularization of P2BE to improve robustness against adversarial perturbations.

\begin{table}[t]
  \centering
  \caption{The test error against LS-PGA white-box attack on CIFAR-10 and CIFAR-100 with various $\lambda$. The value in parenthesis is the test error on clean images. $\lambda$ is the coefficient for embedding smoothness loss in Eqn. \ref{eq_adv}.}
  \label{tab:cifar_adv_lambda}
  \begin{tabular}{ccccccc}
  \hline\noalign{\smallskip}
  & & Test Error (\%)\\
  \hline  \noalign{\smallskip}
\multirow{2}{*}{CIFAR-10}  & \texttt{ConTrain} ($\lambda=0.0$)& 59.3 (12.0) \\ 
                           & \texttt{ConTrain} ($\lambda=1.0$) & \textbf{45.3} (15.7)\\ 
                          \cline{2-2} \noalign{\smallskip}
\multirow{2}{*}{CIFAR-100}  & \texttt{ConTrain} ($\lambda=0.0$)& 72.5 (43.4) \\ 
                            & \texttt{ConTrain} ($\lambda=0.1$) & \textbf{71.5} (42.2)\\ 
  \hline
  \end{tabular}
\end{table}

\noindent \textbf{Does P2BE Require Longer Training?}
We investigate the relationship between the training length and the performances on CIFAR-100-C.
The results are shown in Table \ref{tab:cifar_c_epoch}.
As can be seen in Table \ref{tab:cifar_c_epoch}, it tends to show better performance with the longer training regardless of input space.
Since binary embeddings are learnable in P2BE, the input space is being changed during training, and we have expected that P2BE requires the model to train longer for convergence. However, P2BE shows comparable or better performances with short training periods on the CIFAR-100-C dataset. 

\begin{table}[t]
  \centering
  \caption{The test error of CIFAR-100-C with different epochs for training. The hyperparameters are all same as in Sec \ref{sec:impl} except for epochs.}
  \label{tab:cifar_c_epoch}
  \begin{tabular}{ccccccc}
  \hline\noalign{\smallskip}
 \scriptsize{Epochs} & \scriptsize{Network} & \scriptsize{RGB} & \scriptsize{One-hot}& \scriptsize{Thermometer}& \scriptsize{P2BE (ours)}\\ \hline \noalign{\smallskip}
 & \small{WideResNet} & 36.2 & 36.3 & 36.3 & \textbf{35.9}\\
 50 & \small{DenseNet} &  \textbf{39.6} & 40.1 & 39.7 & 40.3\\ 
 & \small{ResNext} & 34.6 & 34.1 & \textbf{33.8} &  33.9\\ 
                            \cline{2-2} \cline{1-1} \noalign{\smallskip}
& \small{WideResNet} & 34.9 & 34.5  & 35.1 &  \textbf{34.4}\\
100 & \small{DenseNet} & 37.6 & \textbf{37.5} &  37.9& 38.1\\ 
 & \small{ResNext} & 33.5 &  \textbf{33.1} &34.3 &  33.6\\ 
                            \cline{2-2} \cline{1-1} \noalign{\smallskip}
& \small{WideResNet} & 35.4 &  34.4 & 35.1&  \textbf{34.2} \\
200 & \small{DenseNet} &  37.5 & 36.6 & 36.9 &  \textbf{36.3}\\ 
 & \small{ResNext} & 34.2 & 33.2 & 33.9 & \textbf{32.5}\\ 
  \hline
  \end{tabular}
\end{table}

\noindent \textbf{Transferability of Learned Binary Embedding in P2BE.}
It is known that ImageNet-1k pre-trained classification models have good features that are transferable to other tasks.
Then, it is reasonable to consider whether the ImageNet-1k trained P2BE is transferable to other tasks or not.

In this work, we verify the transferability of ImageNet-1k trained P2BE by using CIFAR-10, CIFAR-10-C, CIFAR-100, and CIFAR-100-C.
We show the results in Table \ref{tab:transferability} and the fixed ImageNet-1k trained P2BE outperforms the result with P2BE.
This result indicates that we may be able to get better binary embeddings by using more complex and large-scale datasets.

\begin{table}[tbp]
  \centering
  \caption{The test errors of CIFAR and CIFAR-C datasets with P2BE and fixed ImageNet-1k trained P2BE on WideResNet. Fixed P2BE represents the case that the embedding of P2BE is fixed during the training of the classification model with Eqn. \ref{eq_c}.}
  \label{tab:transferability}
  \vspace{0.2cm}
  \begin{tabular}{ccccc}
  \hline\noalign{\smallskip}
  & \small{Embeddings} & Test Error (\%)\\
  \hline  \noalign{\smallskip}
  \multirow{2}{*}{CIFAR-10} & P2BE & 4.8\\
                            & Fixed P2BE (ImageNet-1k) & \textbf{4.7} \\ \cline{2-2} \noalign{\smallskip}
  \multirow{2}{*}{CIFAR-10-C} & P2BE & 10.3\\
                            & Fixed P2BE (ImageNet-1k) & \textbf{10.1}\\ \cline{2-2} \noalign{\smallskip}
  \multirow{2}{*}{CIFAR-100} & P2BE & 23.2\\
                            & Fixed P2BE (ImageNet-1k) & \textbf{22.8}\\ \cline{2-2} \noalign{\smallskip}
  \multirow{2}{*}{CIFAR-100-C} & P2BE & 34.2\\
                            & Fixed P2BE (ImageNet-1k) & \textbf{33.9} \\
  \hline
  \end{tabular}
\end{table}

\section{Conclusion}

We propose Pixel to Binary Embedding (P2BE) for improving the robustness of CNNs.
P2BE is a learnable binary embedding method as opposed to hand-coded binary embedding methods (e.g., one-hot and thermometer encoding).
We show that P2BE achieves the best robustness performances against adversarial perturbations and common visual corruptions among other hand-coded binary embedding methods.

\section*{Acknowledgement}
These research results were obtained from the commissioned research (No. 225) by National Institute of Information and Communications Technology (NICT), Japan. This work was also supported by Institute of AI and Beyond of the University of Tokyo and JSPS KAKENHI Grant Number JP19H04166.

{\small
\bibliographystyle{IEEEtrans}
\bibliography{egbib}
}

\end{document}